\documentclass[letterpaper,10pt,conference]{ieeeconf}
\IEEEoverridecommandlockouts
\usepackage[T1]{fontenc}
\usepackage{newtxtext}
\usepackage{amsmath,amssymb}
\usepackage{newtxmath}
\usepackage{graphicx,booktabs,multirow,array,adjustbox}
\usepackage{algorithm,algorithmic}
\usepackage{microtype,cite,placeins}
\usepackage{url}
\newcommand{\citep}[1]{\cite{#1}}
\DeclareMathOperator{\Var}{Var}
\DeclareMathOperator{\Cov}{Cov}
\DeclareMathSizes{10}{10}{7}{5}
\DeclareMathSizes{9}{9}{6.3}{5}
\DeclareMathSizes{8}{8}{6}{5}
\usepackage[hidelinks]{hyperref}
\title{HybridFlow: A 2-NFE Generative Policy for Real-Time Robotic Manipulation}
\author{Zhenchen Dong$^{1,*}$, Fulin Chen$^{2,*}$, Jinna Fu$^{3}$, Jiaming Wu$^{4}$\\Qingran Wu$^{5}$, Shengyuan Yu$^{6}$, Hongyu Yu$^{1,\dagger}$, Yide Liu$^{3,\dagger}$
\thanks{$^{*}$Equal contribution. $^{\dagger}$Corresponding authors: Hongyu Yu and Yide Liu.}
\thanks{$^{1}$The Hong Kong Polytechnic University.}
\thanks{$^{2}$Shanghai University of Engineering Science.}
\thanks{$^{3}$Zhejiang University.}
\thanks{$^{4}$Universit\'e Paris-Saclay.}
\thanks{$^{5}$University of California San Diego.}
\thanks{$^{6}$The Chinese University of Hong Kong.}
\thanks{\raggedright Correspondence: \texttt{hongyu.yu@polyu.edu.hk}; \texttt{yide\_liu@zju.edu.cn}.}
}

\begin{document}
\maketitle
\begin{abstract}
Generative policies for robotic manipulation must balance action accuracy with inference latency. We present HybridFlow, a three-stage policy inference procedure requiring two network function evaluations (2-NFE). A Global Jump uses the MeanFlow average velocity to generate a coarse action trajectory; a parameter-free ReNoise interpolation constructs a state at a nonzero refinement time; and a Local Refine queries the instantaneous-velocity limit of the same network at that time. This construction reuses a unified model without distillation. Our analysis characterizes interval-composition errors and the attenuation of endpoint error under ReNoise interpolation. Controlled RoboMimic ablations support the MeanFlow proposal and intermediate-state construction, achieving 95\% average success with reused noise and 95.5\% with fresh noise, versus 78\% for one-step MeanFlow. Across five real-robot settings with all policies running on the same Jetson AGX Thor, HybridFlow improves normalized task performance over 16-step Diffusion Policy by 13--68 points with approximately eightfold lower action-generation latency. Additional experiments demonstrate its compatibility as an action expert in a vision-language-action framework. Project page: \url{https://hybridflow-anonymous.pages.dev/}.
\end{abstract}

\section{Introduction}
\label{sec:intro}
Deploying generative policies on physical robots requires both accurate actions and timely responses. Diffusion Policy~\cite{chi2023diffusion,dp3_2023} generates action sequences through iterative denoising, typically using 8--32 evaluations. Flow-based policies~\cite{pi0_2024,pi05_2025,rdt2} provide an alternative transport formulation, but commonly retain several inference steps. In reactive or dynamic manipulation, the scene can change substantially while a policy computes its next action chunk.

MeanFlow~\cite{meanflow2024,geng2025improvedmeanflowschallenges} offers a different route: instead of integrating an instantaneous velocity field over many short intervals, it learns the average velocity over an interval and can traverse the entire noise-to-action path in one evaluation. This makes it attractive for low-latency control. However, a fast endpoint prediction may lack the precision needed for grasping or contact, and dividing an approximate MeanFlow map into more intervals does not necessarily improve its output.

Our experiments examine how a second network evaluation can improve an already useful one-step policy. In the controlled RoboMimic ablation, one-step MeanFlow achieves 78\% average success and HybridFlow reaches 95\%. Uniform MeanFlow iteration gives 78\%, 72\%, and 60\% at 2, 4, and 16 NFE. These results motivate using the second evaluation for complementary correction. Physical experiments assess the resulting policy under reactive target interception and dynamic grasping.

We address the inference problem using a property already present in the MeanFlow model: its average velocity becomes the instantaneous velocity when the two conditioning times coincide. HybridFlow assigns two complementary roles to one network. A \emph{Global Jump} predicts a coarse endpoint using the average velocity; \emph{ReNoise} interpolates this endpoint with Gaussian noise; and a \emph{Local Refine} queries the diagonal instantaneous field at the matching intermediate time. ReNoise requires no network evaluation, leaving a total budget of 2-NFE. Both reused and independently redrawn noise are effective in our controlled comparison.

Our contributions are a unified 2-NFE inference procedure, an analysis of its refinement inputs, and controlled evaluation of its components. The analysis relates interval-prediction errors to generated queries and characterizes ReNoise for either noise source. Ablations examine proposal choice, intermediate-state construction, and noise-source choice. Simulation, five physical evaluation settings, and VLA action-expert experiments assess the resulting balance of accuracy and latency.

\section{Related Work}
\label{sec:related}
\textbf{Generative manipulation policies.}
Diffusion Policy~\cite{chi2023diffusion} models conditional action sequences by denoising. Extensions use 3D representations~\cite{dp3_2023}, hierarchical generation~\cite{lu2025h3dp}, and visual-tactile observations~\cite{tian2026vitas}. Flow matching~\cite{lipman2023flow} instead learns a transport field, with Rectified Flow~\cite{liu2023rectified} using straight interpolation paths. MeanFlow learns interval-average velocities~\cite{meanflow2024}; applications and extensions include images~\cite{lu2026pixelmeanflow,huang2026rmflow}, PDEs~\cite{shen2026twostepdiffusion}, and audio~\cite{shimizu2025meanflowtse}. Learning instantaneous and average fields can interact: Kim et al.~\cite{kim2026understanding} find that large-gap average-velocity learning depends on an established instantaneous field and can interfere with it. This motivates examining both training and inference before attributing poor control to a fundamental limitation.

\textbf{MeanFlow policies.}
MP1~\cite{mp1_meanflow} combines point-cloud conditioning, classifier-free guidance, and dispersive regularization. DM1~\cite{dm1} also regularizes representations and reports vanilla MeanFlow simulation baselines. OMP~\cite{fang2026omp} adds directional alignment and evaluates ablations removing its added losses. MeanFlow-based One-Step VLA~\cite{chen2026meanflowvla} reports successful one-step physical manipulation alongside sensitivity to training hyperparameters and action-chunk length. These studies establish successful applications under their respective observation, training, and control settings. Our evaluation examines a complementary inference design: using a second query to refine an interval-average proposal within one model. We assess this design through controlled component ablations and UMI physical tasks, with performance interpreted relative to the evaluated configurations.

\textbf{Few-step and refinement-based policies.}
Consistency Policy~\cite{prasad2024consistency}, ManiCM~\cite{chen2024manicm}, and OneDP~\cite{wang2025onedp} accelerate generation through distillation or consistency learning. ShortCut~\cite{frans2024shortcut} conditions on step size; OFP~\cite{li2026ofp} uses self-distillation; and DMPO~\cite{zou2026dmpo} extends MeanFlow with policy optimization. Energy-based single-pass prediction provides another direction~\cite{jia2025energy}. These approaches modify the training procedure to support fewer inference steps.

RMFlow~\cite{huang2026rmflow} augments MeanFlow training with a likelihood objective and injects independent noise into its generated output, retaining 1-NFE. HybridFlow instead constructs an intermediate state and evaluates the same network again for instantaneous-velocity correction. STEP~\cite{li2026step} uses an auxiliary spatiotemporal predictor to initialize diffusion refinement and velocity-aware perturbation to mitigate execution stalls. HybridFlow obtains its proposal from the MeanFlow network itself. MIP~\cite{pan2026mip} trains a two-step regressor with intermediate supervision and training-time noise, then performs deterministic inference. HybridFlow retains MeanFlow training and combines interval-average and diagonal velocity queries during inference. These methods share the idea of refinement, but differ in their training objectives, proposal construction, and use of additional network evaluations.

\section{Method}
\label{sec:method}
\subsection{A Unified Average- and Instantaneous-Velocity Model}
We learn a visuomotor policy $p(x\mid c)$ over action chunks $x\in\mathbb{R}^d$, where $c=f_\phi(o)$ encodes observations. Rectified Flow transports Gaussian noise at $t=1$ toward actions at $t=0$ through an instantaneous field $v(z_t,t,c)$. MeanFlow defines the interval-average velocity
\begin{equation}
u(z_t,r,t,c)=\frac{1}{t-r}\int_r^t v(z_\tau,\tau,c)\,d\tau,
\label{eq:avg_vel}
\end{equation}
and the MeanFlow identity gives
\begin{equation}
u(z_t,r,t,c)=v(z_t,t,c)-(t-r)\frac{d}{dt}u(z_t,r,t,c).
\label{eq:mf_identity}
\end{equation}
At $r=t$, the interval-average field reduces to the instantaneous field: $u(z_t,t,t,c)=v(z_t,t,c)$. Thus, one network $u_\theta$ can support long-interval transport and local velocity queries.

\begin{figure}[!t]
\centering
\includegraphics[width=\linewidth]{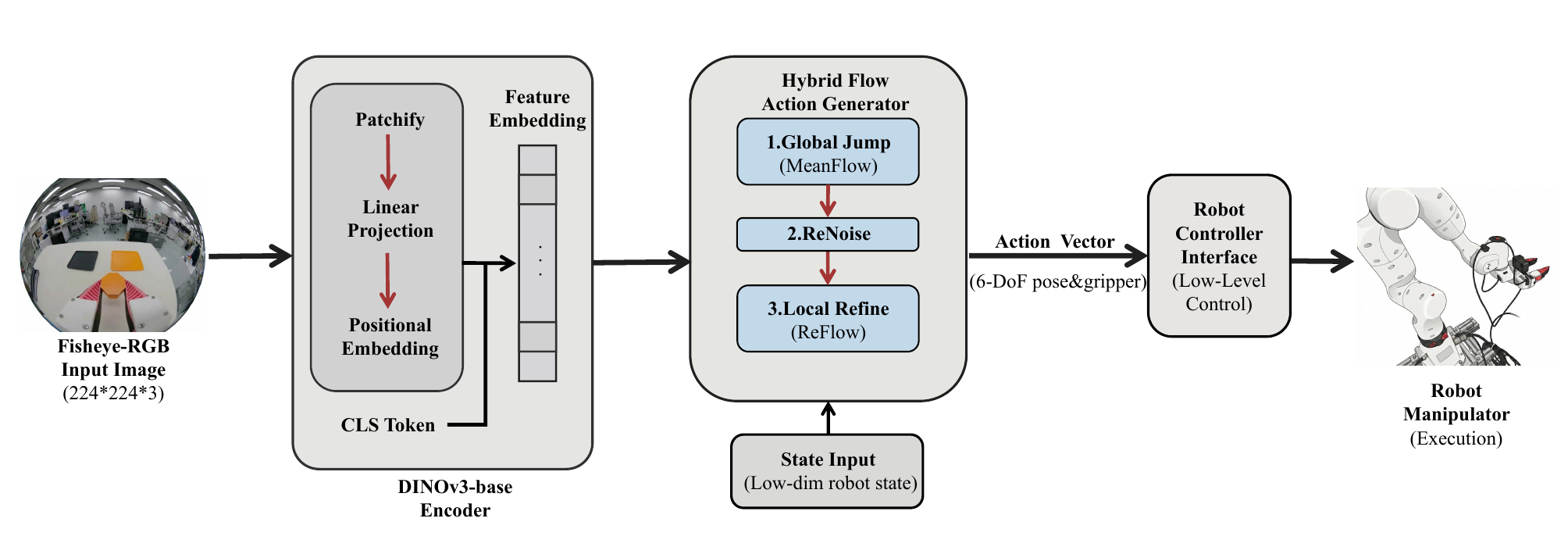}
\caption{\textbf{Real-robot system architecture.} Fisheye RGB observations are encoded by a pretrained vision encoder, fine-tuned with the policy. Visual features and low-dimensional robot state condition the HybridFlow generator, which produces an action trajectory for the controller.}
\label{fig:architecture}
\end{figure}

\begin{figure*}[!t]
\centering
\includegraphics[width=0.86\textwidth]{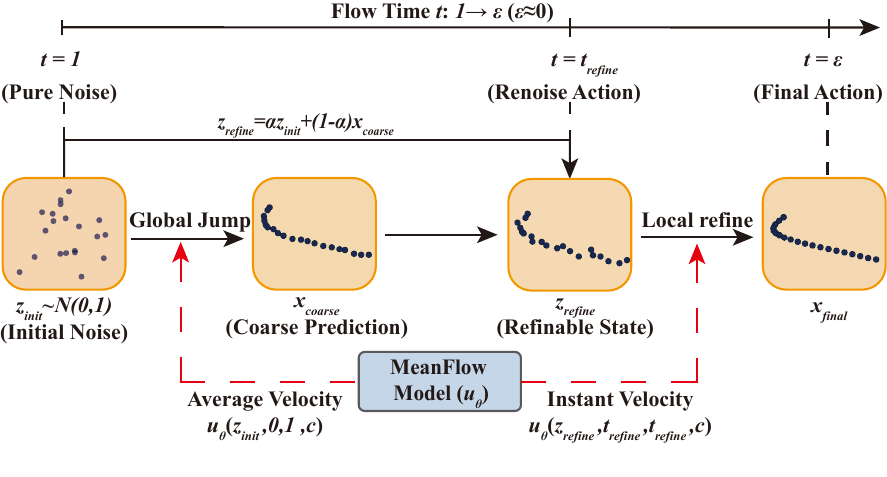}
\caption{\textbf{HybridFlow: three stages, two network evaluations.} Global Jump predicts an endpoint at $t=0$; ReNoise returns to $t_{\mathrm{refine}}=\alpha$; Local Refine advances to $t=\epsilon\approx0$. The stages show computation order, with reused noise as the default. Only the first and third stages evaluate the network.}
\label{fig:principle}
\end{figure*}

Training follows the MeanFlow formulation with $(r,t)$ conditioning and a Jacobian-vector product for the total derivative in Eq.~\eqref{eq:mf_identity}. Half of each batch uses $r=t$, where the derivative term vanishes and the objective reduces to instantaneous-velocity learning; the remaining half uses distinct times to learn interval-average transport. The same model is used throughout inference, with no separate refinement network or distillation stage. Figure~\ref{fig:architecture} shows the observation-to-control pipeline.

\subsection{Interval Composition and Query Shift}
\label{sec:query_shift}
For $t_k=1-k/K$ and $\Delta t=1/K$, a uniform $K$-step sampler applies
\begin{equation}
z^{(k+1)}=z^{(k)}-\Delta t\,u_\theta(z^{(k)},t_{k+1},t_k,c).
\label{eq:mf_steps}
\end{equation}
Exact transport maps compose consistently across intervals. MeanFlow training on interpolated queries $(z_t,r,t)$ does not explicitly enforce this consistency on the model-generated states encountered during iterative inference.

We refer to the resulting change in state marginals and their coupling with time endpoints as \emph{joint-query distribution shift}. For a fixed query, writing $\widehat u=u+\delta$ yields $\widehat z_r=z_r-(t-r)\delta$. Prediction error therefore perturbs the next query even under unguided inference. Iterative accuracy depends on both the interval approximation and the states encountered during composition.

Let $z_*^{(k)}$ follow the exact interval maps on the same schedule and from the same initial noise as $z^{(k)}$. Set $E_k=\|z^{(k)}-z_*^{(k)}\|$ and $\Delta t=1/K$. For the generated query $q_k=(z^{(k)},t_{k+1},t_k,c)$, define the approximation error by
\begin{equation}
e_k=u_\theta(q_k)-u^*(q_k).
\label{eq:app_error}
\end{equation}
Assume $u^*$ is $L$-Lipschitz in state along the compared trajectories~\cite{chen2023probability_flow}. Adding and subtracting its value at the generated state, then applying the triangle inequality, gives
\begin{equation}
E_{k+1}\leq (1+L\Delta t)E_k+\Delta t\|e_k\|.
\label{eq:app_recursion}
\end{equation}
Since $E_0=0$, unrolling the recursion yields
\begin{equation}
E_K\leq\Delta t\sum_{k=0}^{K-1}(1+L\Delta t)^{K-1-k}\|e_k\|.
\label{eq:recursion}
\end{equation}
For $L>0$ and a uniform error bound $\|e_k\|\leq\bar e$, this further implies
\begin{equation}
E_K\leq\frac{\bar e}{L}\left[(1+L/K)^K-1\right].
\label{eq:app_uniform}
\end{equation}
As $K\to\infty$, the bound approaches $\bar e(e^L-1)/L$; for $L=0$, it reduces to $E_K\leq\bar e$. This is fixed-interval amplification, not exponential growth with NFE. Local sensitivity and query shift can affect prediction errors~\cite{mooney2024lipschitz_score,zhou2025flow_error}. The bound is general; Table~\ref{tab:ablation} measures the NFE dependence of the evaluated MeanFlow model.

Our RoboMimic experiments measure this dependence through task success (Section~\ref{sec:diagnostics}). A related image-generation study reports CIFAR-10 FID $2.80/2.84/5.14/4.34$ at $1/2/4/8$ MeanFlow steps~\cite{ai2025joint}. These observations motivate a second query designed for local correction.

\subsection{Global Jump, ReNoise, and Local Refine}
\label{sec:hybridflow}
HybridFlow uses the two query types for distinct purposes (Fig.~\ref{fig:principle}). We write $z_1=z_{\mathrm{init}}$, $x_c=x_{\mathrm{coarse}}$, and $z_\alpha=z_{\mathrm{refine}}$, with $t_{\mathrm{refine}}=\alpha$. The final query advances to the near-zero terminal time $\epsilon$ shown in the figure; $\epsilon=0$ recovers the ideal data endpoint used in the analysis.

\textbf{Global Jump.} Sample $z_1\sim\mathcal N(0,I)$ and use the interval-average field to predict an endpoint:
\begin{equation}
x_c=z_1-u_\theta(z_1,0,1,c).
\label{eq:global_jump}
\end{equation}
The interval length is one, so the predicted average velocity represents the full-interval displacement.

\textbf{ReNoise.} Let $\eta=z_1$ for reused noise (default), or let $\eta=\xi$ with $\xi\sim\mathcal N(0,I)$ drawn independently of the proposal for the fresh-noise variant. Form
\begin{equation}
z_\alpha=\alpha\eta+(1-\alpha)x_c,\qquad 0<\alpha<1.
\label{eq:renoise}
\end{equation}
The query time is set to the interpolation coefficient, $t=\alpha$, giving a time-aligned refinement state without another network evaluation. For a reference endpoint $x$, use the same draw $\eta$ to define $z_\alpha^*=\alpha\eta+(1-\alpha)x$. With $e=x_c-x$,
\begin{equation}
z_\alpha-z_\alpha^*=(1-\alpha)e.
\label{eq:renoise_error}
\end{equation}
This identity holds for both noise sources: the endpoint error is scaled by $1-\alpha$ relative to a reference using the same $\eta$. It does not require reusing the Global Jump noise or guarantee a match to the training marginal. Noise-source dependence enters the covariance analysis below.

\textbf{Local Refine.} Query the same network on the diagonal $r=t=\alpha$ and take one instantaneous-velocity step to $0\leq\epsilon<\alpha$:
\begin{equation}
\widehat x=z_\alpha-(\alpha-\epsilon)u_\theta(z_\alpha,\alpha,\alpha,c).
\label{eq:local_refine}
\end{equation}
Here $u_\theta(z_\alpha,\alpha,\alpha,c)$ estimates $v(z_\alpha,\alpha,c)$; multiplying by $\alpha-\epsilon$ gives the displacement to the terminal time. The output $\widehat x=x_{\mathrm{final}}$ uses two evaluations in total (Algorithm~\ref{alg:hybridflow}).

\begin{algorithm}[!htbp]
\caption{HybridFlow inference (3-stage, 2-NFE)}
\label{alg:hybridflow}
\begin{algorithmic}[1]
\REQUIRE Model $u_\theta$, encoding $c$, ratio $\alpha$, terminal $\epsilon\approx0$
\STATE Sample $z_1\sim\mathcal N(0,I)$
\STATE $x_c\leftarrow z_1-u_\theta(z_1,0,1,c)$ \COMMENT{Global Jump}
\STATE Set $\eta\leftarrow z_1$ (default) or draw independent $\eta\sim\mathcal N(0,I)$
\STATE $z_\alpha\leftarrow\alpha\eta+(1-\alpha)x_c$ \COMMENT{ReNoise}
\STATE $v_\alpha\leftarrow u_\theta(z_\alpha,\alpha,\alpha,c)$ \COMMENT{Local Refine}
\STATE $\widehat x\leftarrow z_\alpha-(\alpha-\epsilon)v_\alpha$
\RETURN $\widehat x$
\end{algorithmic}
\end{algorithm}

\textbf{Two-step inference variants.} Two-step ReFlow uses two instantaneous-velocity Euler steps. ReFlow--ReNoise--ReFlow replaces HybridFlow's interval-average proposal while retaining its correction pipeline; MeanFlow--ReFlow without ReNoise omits the interpolation. Section~\ref{sec:ablations} compares these variants.

\subsection{Noise Source and Refinement-Time Alignment}
\label{sec:variance}
For fixed observation conditioning and any action coordinate, write $V_\eta=\Var(\eta)$, $V_c=\Var(x_c)$, and $C_{\eta c}=\Cov(\eta,x_c)$. Then
\begin{equation}
\Var(z_\alpha)=\alpha^2 V_\eta+(1-\alpha)^2 V_c+2\alpha(1-\alpha)C_{\eta c}.
\label{eq:variance_exact}
\end{equation}
Fresh noise is independent of $x_c$, giving $C_{\eta c}=0$. With reused noise, $x_c$ depends on $z_1$, so the covariance is generally nonzero. For unit noise variance, negligible covariance, and small action variance,
\begin{equation}
\Var(z_\alpha)\approx\alpha^2 V_\eta+(1-\alpha)^2 V_c\approx\alpha^2.
\label{eq:variance_heuristic}
\end{equation}
This approximation motivates matching interpolation strength to query time, as in noise-level conditioning~\cite{ho2020ddpm}; exact marginal matching also depends on the endpoint distribution. Equation~\eqref{eq:renoise_error} requires none of these variance assumptions. The two tested sources give comparable task-level averages (Section~\ref{sec:ablations}), supporting effective refinement with either noise source in this evaluation. We select $\alpha$ empirically (Fig.~\ref{fig:alpha_ablation}).

\section{Experiments}
\label{sec:experiments}
We evaluate four RoboMimic tasks~\cite{robomimic2021}---Can, Lift, Square, and Transport---and five physical settings: in-distribution pick-and-place (PP-ID), unseen-color pick-and-place (Pink (OOD)), eyepatch folding (EP), whack-a-mole (WM), and running-car transport (CT). Table~\ref{tab:combined_results} consolidates simulation and real-robot results. We first examine training and inference ablations, then assess task performance and action-expert transfer.

\begin{table*}[!tp]
\centering
\caption{\textbf{Simulation and real-robot results.} Simulation success rate and normalized real-robot task performance (\%).}
\label{tab:combined_results}
\label{tab:simulation_results}
\label{tab:real_robot}
\small
\setlength{\tabcolsep}{3pt}
\begin{minipage}{\textwidth}
\textbf{(a)} \textbf{Simulation results on image-based RoboMimic.} Success rate (\%) at selected NFE budgets. Best at each (task, NFE) in \textbf{bold}. HybridFlow uses a fixed 2-NFE.
\end{minipage}\par\smallskip
\begin{adjustbox}{max width=\textwidth}
\begin{tabular}{lc cccc cccc cccc cccc}
\toprule
& & \multicolumn{4}{c}{\textbf{Can}} & \multicolumn{4}{c}{\textbf{Lift}} & \multicolumn{4}{c}{\textbf{Square}} & \multicolumn{4}{c}{\textbf{Transport}} \\
\cmidrule(lr){3-6} \cmidrule(lr){7-10} \cmidrule(lr){11-14} \cmidrule(lr){15-18}
\textbf{Policy} & \textbf{Steps} & 1 & 2 & 4 & 16 & 1 & 2 & 4 & 16 & 1 & 2 & 4 & 16 & 1 & 2 & 4 & 16 \\
\midrule
DDIM~\citep{chi2023diffusion} & 1--16 & 84 & 94 & \textbf{100} & \textbf{100} & \textbf{100} & \textbf{100} & \textbf{100} & 98 & 72 & 74 & 80 & 76 & 74 & 78 & 84 & \textbf{88} \\
ReFlow~\citep{liu2023rectified} & 1--16 & 58 & 52 & 22 & 52 & 42 & 84 & 90 & 76 & 52 & 48 & 56 & 60 & 58 & 48 & 44 & 70 \\
ShortCut~\citep{frans2024shortcut} & 1--16 & 54 & 66 & 62 & 16 & 30 & 76 & 92 & 86 & 40 & 52 & 62 & 74 & 50 & 54 & 56 & 64 \\
OFP~\citep{li2026ofp} & 1 & 82 & -- & -- & -- & 90 & -- & -- & -- & 78 & -- & -- & -- & 72 & -- & -- & -- \\
OneDP~\citep{wang2025onedp} & 1 & 78 & -- & -- & -- & 92 & -- & -- & -- & 84 & -- & -- & -- & 64 & -- & -- & -- \\
DMPO~\citep{zou2026dmpo} & 1 & \textbf{100} & -- & -- & -- & \textbf{100} & -- & -- & -- & 83 & -- & -- & -- & \textbf{88} & -- & -- & -- \\
STEP~\citep{li2026step} & 2--4 & -- & \textbf{100} & \textbf{100} & -- & -- & \textbf{100} & \textbf{100} & -- & -- & 96 & 92 & -- & -- & 86 & \textbf{88} & -- \\
\midrule
\textbf{Ours} & 2 & -- & \textbf{100} & -- & -- & -- & \textbf{100} & -- & -- & -- & \textbf{98} & -- & -- & -- & 82 & -- & -- \\
\bottomrule
\end{tabular}
\end{adjustbox}
\par\medskip
\begin{minipage}{\textwidth}
\textbf{(b)} \textbf{Real robot results: all methods deployed on the same Jetson AGX Thor.} Normalized task performance (\%) across 5 settings, with the same observation encoder, action horizon, robot controller, training data, and comparable action-generator backbone. M-Avg.\ is the mean of the displayed task scores.
\end{minipage}\par\smallskip
\setlength{\tabcolsep}{6pt}
\begin{adjustbox}{max width=\textwidth}
\begin{tabular}{lcccccccc}
\toprule
\textbf{Policy} & \textbf{Steps} & \textbf{T\textsubscript{inf}} & \textbf{PP-ID} & \textbf{Pink (OOD)} & \textbf{EP} & \textbf{WM} & \textbf{CT} & \textbf{M-Avg.} \\
\midrule
MeanFlow & 1 & $\sim$10ms & 0.0 & 0.0 & 0.0 & 0.0 & 0.0 & 0.0 \\
DDIM~\citep{chi2023diffusion} & 2 & $\sim$19ms & 0.0 & 0.0 & 2.5 & 0.0 & 0.0 & 0.5 \\
ReFlow~\citep{liu2023rectified} & 2 & $\sim$19ms & 10.0 & 7.5 & 5.0 & 2.3 & 0.0 & 5.0 \\
DDIM~\citep{chi2023diffusion} & 16 & $\sim$152ms & 63.8 & 47.5 & 51.3 & 47.7 & 0.0 & 42.1 \\
\midrule
\textbf{Ours} & \textbf{2} & $\sim$\textbf{19ms} & \textbf{86.3} & \textbf{68.8} & \textbf{66.3} & \textbf{60.7} & \textbf{68.3} & \textbf{70.1} \\
\bottomrule
\end{tabular}
\end{adjustbox}
\par\smallskip
\parbox{\textwidth}{\scriptsize \textbf{PP-ID}: Pick-and-Place In-Distribution; \textbf{Pink (OOD)}: unseen-color pick-and-place only; \textbf{EP}: Eyepatch Folding; \textbf{WM}: Whack-a-Mole (normalized aggregate score); \textbf{CT}: Running Car Transport (HybridFlow: 43/63; each baseline: 0/63). \textbf{T\textsubscript{inf}}: action-generator inference latency on Jetson AGX Thor, excluding camera acquisition, visual encoding, and robot execution.}
\end{table*}

\subsection{Simulation Protocol and MeanFlow Diagnostics}
\label{sec:diagnostics}
The simulation evaluation uses image-based RoboMimic tasks, a ViT observation encoder, RGB observations from three cameras, and 6-DoF-plus-gripper commands, with 100 episodes per task. Baselines include DDIM~\cite{chi2023diffusion}, ReFlow~\cite{liu2023rectified}, ShortCut~\cite{frans2024shortcut}, OFP~\cite{li2026ofp}, OneDP~\cite{wang2025onedp}, DMPO~\cite{zou2026dmpo}, and STEP~\cite{li2026step}. These cover iterative diffusion/flow, few-step training, and warm-start refinement. Table~\ref{tab:ablation} additionally evaluates plain MeanFlow and the inference variants of the unified model. Implementation settings are collected in Section~\ref{sec:implementation}.

\textbf{Training and time sampling.}
Simulation training uses 2,000 epochs. Both logit-normal and uniform time sampling reach validation loss near 0.005; RoboMimic success is 70\% with uniform sampling and 78\% with logit-normal sampling (default). Changing the time-sampling distribution therefore does not improve MeanFlow success in this comparison. We use within-method validation loss to monitor training and task success to compare control quality across regression targets. The separate real-robot training-duration study is described in Section~\ref{sec:real}.

\subsection{Controlled Inference Ablations}
\label{sec:ablations}
The controlled comparison uses one unified checkpoint and 100 episodes per task. Here, ReFlow-mode inference denotes diagonal queries of that model. Plain MeanFlow and ReFlow-mode inference use the uniform schedule $1\to0.5\to0$ for two-step sampling. The ReFlow--ReNoise--ReFlow variant replaces the full-interval MeanFlow proposal with an instantaneous-field endpoint prediction and retains the same interpolation and final correction. The no-ReNoise variant retains the MeanFlow proposal and final ReFlow correction but omits interpolation.

\begin{table}[!htbp]
\centering
\caption{\textbf{Controlled RoboMimic inference ablation.} Average success (\%) over the four tasks. RN denotes ReNoise. All variants query the same unified checkpoint.}
\label{tab:ablation}
\small
\setlength{\tabcolsep}{3pt}
\begin{tabular}{@{}>{\raggedright\arraybackslash}p{0.42\columnwidth}cc@{}}
\toprule
Inference procedure & NFE & Success (\%) \\
\midrule
Plain MeanFlow & 1/2/4/16 & 78/78/72/60 \\
MeanFlow--ReFlow, no RN & 2 & 24 \\
Two-step ReFlow & 2 & 58 \\
ReFlow--RN--ReFlow & 2 & 82 \\
HybridFlow (same noise) & 2 & 95 \\
HybridFlow (fresh noise) & 2 & \textbf{95.5} \\
\bottomrule
\end{tabular}
\end{table}

Table~\ref{tab:ablation} examines proposal choice, intermediate-state construction, and noise source. Fixing the MeanFlow proposal and diagonal correction, adding ReNoise raises success from 24\% to 95\%. Fixing ReNoise and correction, replacing the ReFlow proposal with MeanFlow raises success from 82\% to 95\%. These controls support combining the full-interval proposal with a nonzero-time refinement state. The noise-source comparison below tests whether this improvement requires reusing the initial draw. Separately, MeanFlow iteration degrades at larger NFE; the 58\%-to-82\% comparison changes both interpolation and inference path. The no-ReNoise variant also yields approximately zero success in the real-robot component check.

\begin{figure}[!htbp]
\centering
\includegraphics[width=0.94\linewidth]{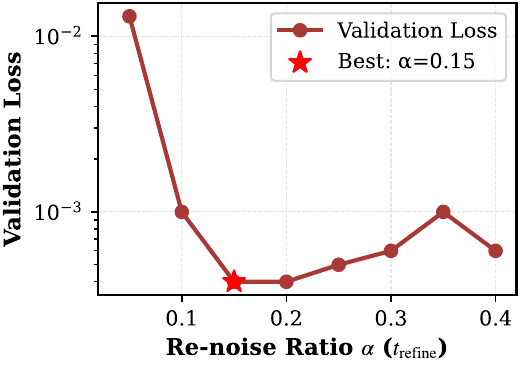}
\caption{\textbf{ReNoise ratio ablation.} The validation curve favors $\alpha\in[0.15,0.20]$. We select $\alpha=0.15$ and use it throughout the evaluated tasks.}
\label{fig:alpha_ablation}
\end{figure}

\begin{figure*}[!t]
\centering
\includegraphics[width=\linewidth]{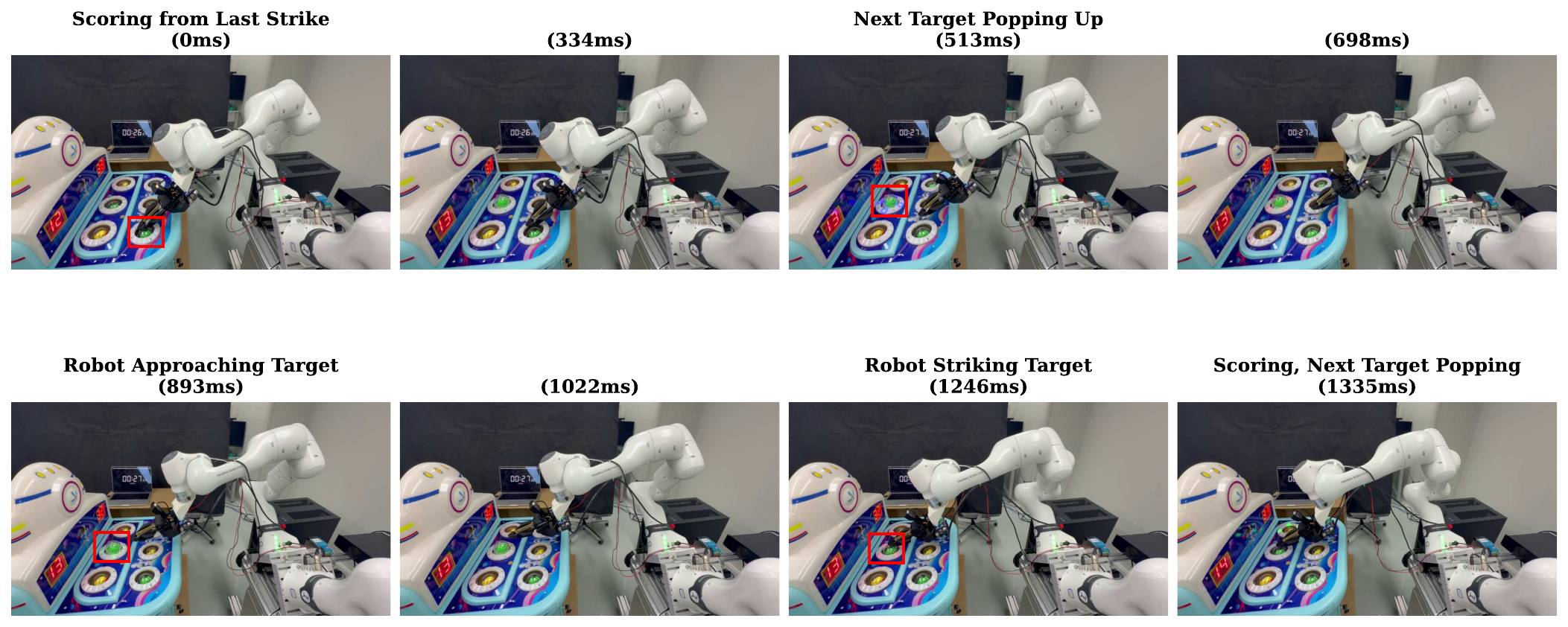}
\caption{\textbf{Whack-a-mole.} The robot responds to randomly activated targets before they retract. WM uses normalized aggregate score over six rounds (50 points each).}
\label{fig:whack_a_mole}
\end{figure*}

\begin{figure*}[!t]
\centering
\includegraphics[width=\linewidth]{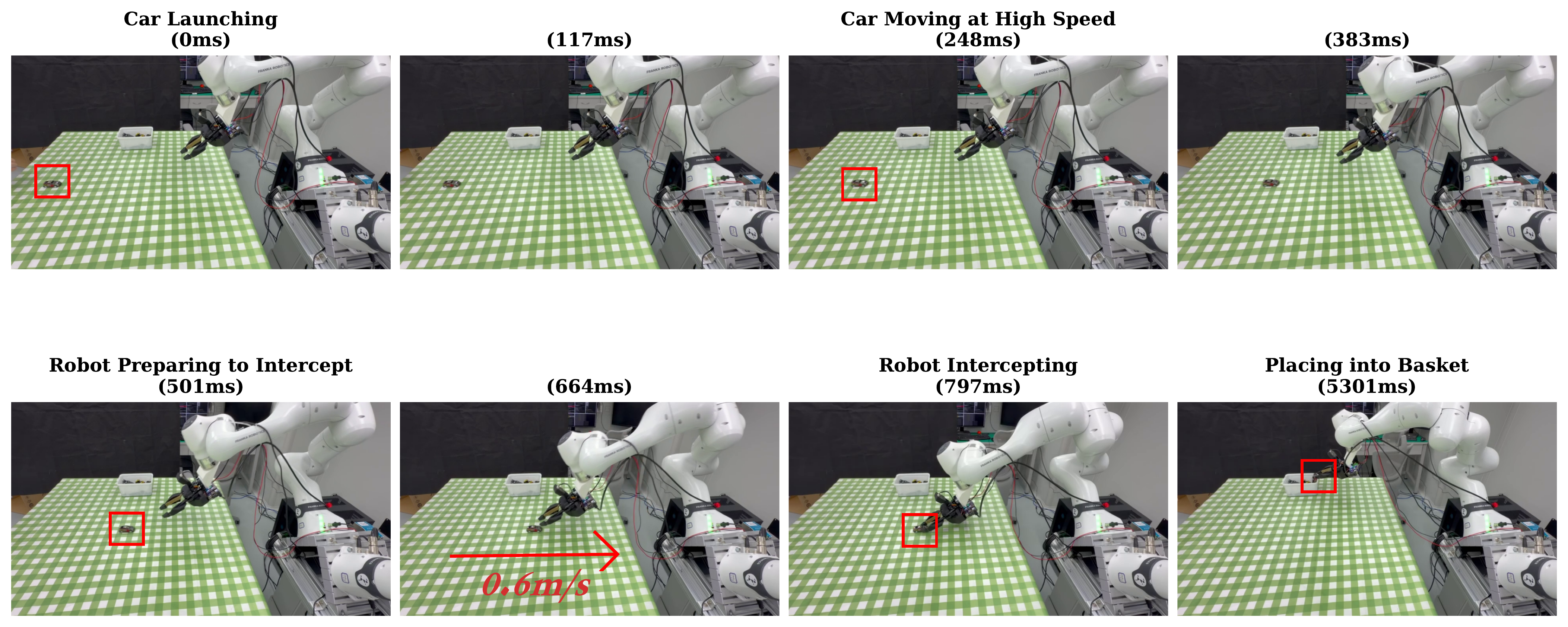}
\caption{\textbf{Running-car transport.} The robot intercepts a car moving at approximately 0.6\,m/s and places it in a basket. HybridFlow succeeds in 43/63 trials (68.3\%); each evaluated baseline achieves 0/63.}
\label{fig:car_transport}
\end{figure*}

\textbf{Noise-source ablation.}
Keeping the checkpoint, refinement ratio, and diagonal correction fixed, we change only $\eta$ in Eq.~\eqref{eq:renoise} from the initial draw $z_1$ to independent $\xi\sim\mathcal{N}(0,I)$. Fresh noise achieves 100\%, 100\%, 96\%, and 86\% on Can, Lift, Square, and Transport, respectively: a 95.5\% average versus 95\% with reused noise. Both sources retain the improvement over the 24\% no-ReNoise variant. The tested refinement is effective with either source; the evidence supports constructing a time-aligned state before the diagonal query, rather than requiring reuse of the original draw.

\textbf{ReNoise strength.}
Figure~\ref{fig:alpha_ablation} reports validation loss as a function of $\alpha$. At $\alpha=0.05$, the loss is 0.013; the minimum near $\alpha\in[0.15,0.20]$ is approximately $4\times10^{-4}$, and larger values above 0.25 degrade quality. We use $\alpha=0.15$ for all evaluated tasks without per-task retuning. This supports the choice within the tested settings, while other action distributions or backbones may require a different value.

\textbf{Comparison with other policies.}
In Table~\ref{tab:simulation_results}(a), HybridFlow reaches 100\% on Can and Lift at 2-NFE. On Square it achieves 98\%, compared with 96\% for 2-NFE STEP, 84\% for OneDP, and 83\% for DMPO. Transport remains challenging: HybridFlow achieves 82\%, versus 86\% for 2-NFE STEP and 88\% for 16-step DDIM. The task-level results place HybridFlow and STEP at comparable performance under the same denoising budget. Their inference costs have different components: STEP includes a separately trained predictor in addition to denoising, while HybridFlow makes two calls to its unified model.

\subsection{Real-Robot Manipulation}
\label{sec:real}

\textbf{Setup and task specification.}
Physical experiments use a Franka Research 3 arm and gripper. \textbf{All real-robot policy inference, including HybridFlow and every baseline, runs locally on the same Jetson AGX Thor.} The UMI setup~\cite{umi2024} uses $224\times224$ fisheye RGB images and low-dimensional robot state (Fig.~\ref{fig:architecture}), with a DINOv3-base encoder~\cite{simeoni2025dinov3} fine-tuned end-to-end, and approximately 300 demonstrations per task. The baselines share training data, observation encoder, action horizon, controller, and the tuning pipeline, with comparable action-generator backbones. Each method is evaluated over 80 trials per setting for PP-ID, Pink (OOD), and EP, 63 trials for CT, and six rounds for WM. Table~\ref{tab:real_robot}(b) reports each evaluation setting and its unweighted macro-average. Bracketed intervals below are 95\% Wilson confidence intervals in percent, assuming independent Bernoulli trials for the binary tasks.

\begin{figure}[!htbp]
\centering
\includegraphics[width=\linewidth]{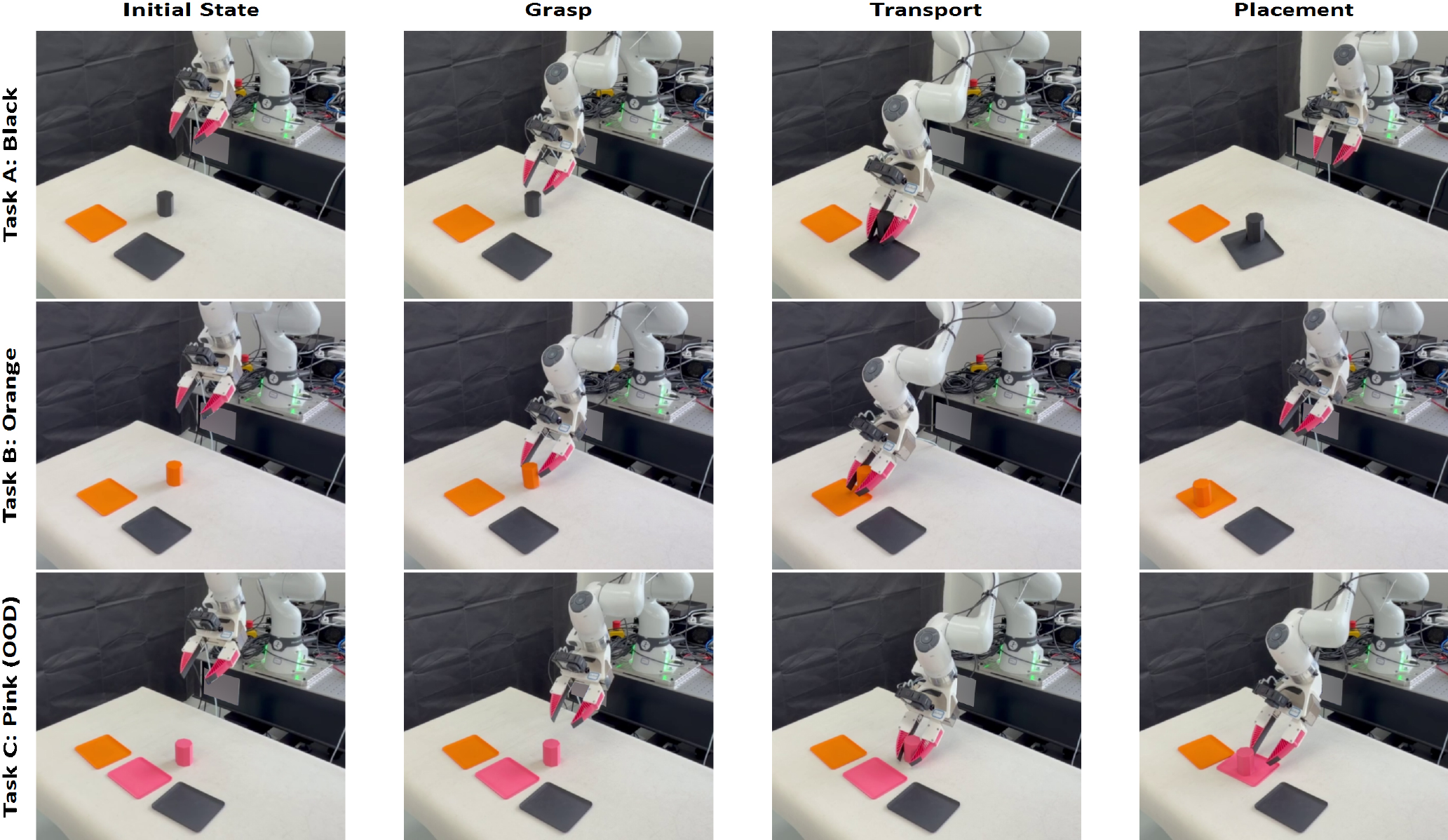}
\caption{\textbf{Pick-and-place execution.} Each row shows initial state, grasp, transport, and placement. Black and Orange appear in training; Pink (OOD) tests unseen-color generalization only.}
\label{fig:real_robot}
\end{figure}

\begin{figure}[!htbp]
\centering
\includegraphics[width=\linewidth]{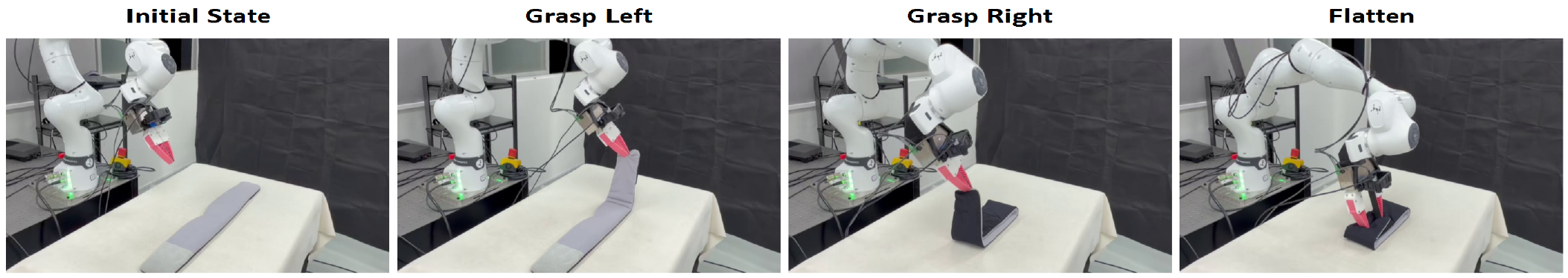}
\caption{\textbf{Eyepatch folding.} The sequence shows initial state, grasping the left and right edges, and folding and flattening the fabric.}
\label{fig:eyepatch}
\end{figure}

Pick-and-place succeeds when the object is placed in its matching tray. Black and Orange are training/ID colors; Pink (OOD) denotes unseen pink objects; OOD refers only to color. The objects are static, while the arm continues executing actions between policy predictions. Eyepatch folding requires successive grasps to fold and flatten the fabric. Figures~\ref{fig:real_robot} and~\ref{fig:eyepatch} show the corresponding execution sequences. HybridFlow improves over 16-step DDIM by 15--22.5 percentage points across these three settings.

On PP-ID, Pink (OOD), and EP, respectively, HybridFlow succeeds in 69/80 [77.0, 92.1], 55/80 [57.9, 77.8], and 53/80 [55.4, 75.7] trials. DDIM-16 achieves 51/80 [52.8, 73.4], 38/80 [36.9, 58.3], and 41/80 [40.5, 61.9]; ReFlow-2 achieves 8/80 [5.2, 18.5], 6/80 [3.5, 15.4], and 4/80 [2.0, 12.2]. MeanFlow records 0/80 [0.0, 4.6] in each setting. DDIM-2 records 0/80 [0.0, 4.6] on both pick-and-place settings and 2/80 [0.7, 8.7] on EP.

\textbf{Reactive control.}
In whack-a-mole, one of six targets activates at random and retracts after 2\,s, with occasional two-target activations outside the demonstration pattern. Each of the six rounds per method has a maximum score of 50 points. Aggregate scores are 0, 0, 7, 143, and 182 out of 300 for MeanFlow, DDIM-2, ReFlow-2, DDIM-16, and HybridFlow, respectively, corresponding to 0\%, 0\%, 2.3\%, 47.7\%, and 60.7\%. WM is evaluated by normalized aggregate score, not binary task success; its points are not treated as independent trials for confidence intervals.

\textbf{Dynamic grasping.}
Running-car transport requires intercepting a toy car moving at approximately 0.6\,m/s and placing it in a basket. HybridFlow succeeds in 43/63 trials (68.3\%; [56.0, 78.4]), while each evaluated baseline achieves 0/63 (0\%; [0.0, 5.7]). At this speed, 152\,ms and 19\,ms of action generation correspond to approximately 9.1\,cm and 1.1\,cm of car motion, respectively, quantifying the timing pressure during interception.

\begin{table*}[!tp]
\centering
\caption{\textbf{VLA action expert transfer.} HybridFlow replaces only the action expert; VLA backbone unchanged.}
\label{tab:vla}
\small
\begin{tabular}{lccccccc}
\toprule
& & \multicolumn{4}{c}{\textbf{LIBERO Benchmark}} & \multicolumn{2}{c}{\textbf{RoboTwin 2.0}} \\
\cmidrule(lr){3-6} \cmidrule(lr){7-8}
\textbf{Method} & \textbf{Steps} & Spatial & Object & Goal & Avg. & Easy/Hard & Avg. \\
\midrule
StarVLA-ReFlow & 4 & 95.6 & 99.6 & 96.0 & 97.1 & 88.2/88.3 & 88.3 \\
StarVLA-HybridFlow & 2 & \textbf{96.2} & 99.4 & \textbf{97.5} & \textbf{97.7} & \textbf{89.6}/\textbf{89.0} & \textbf{89.3} \\
\bottomrule
\end{tabular}
\parbox{\textwidth}{\footnotesize\textit{Note:} LIBERO Avg. is the unweighted mean of Spatial, Object, and Goal. RoboTwin results are averaged over 3 seeds.}
\end{table*}

\begin{table*}[!tp]
\centering
\small
\caption{\textbf{Hyperparameters.} Left: Simulation (RoboMimic). Right: Real robot (UMI).}
\label{tab:hyper}
\begin{minipage}[t]{0.48\textwidth}
\centering
\begin{tabular}{@{}ll@{}}
\toprule
\multicolumn{2}{c}{\textbf{Simulation (RoboMimic)}} \\
\midrule
Network & MLP 768-768-768, Mish \\
Vision & ViT (patch 8, dim 128, 4 heads) \\
Obs & 3 $\times$ 96$\times$96 RGB + 9-d state \\
Action / horizon & 7-d / 4 steps \\
Optimizer & AdamW, lr=$10^{-4}$, wd=$10^{-6}$ \\
Batch / epochs & 256 / 2000 \\
LR sched & Cosine, warmup 100 \\
EMA & 0.995, start epoch 20 \\
Flow ratio & 0.5 \\
$\alpha$ & 0.15 \\
Train / Infer & RTX 5090 \\
\bottomrule
\end{tabular}
\end{minipage}
\hfill
\begin{minipage}[t]{0.48\textwidth}
\centering
\begin{tabular}{@{}ll@{}}
\toprule
\multicolumn{2}{c}{\textbf{Real Robot (UMI)}} \\
\midrule
Network & 1D UNet [256,512,1024] \\
Vision & DINOv3-base, fine-tuned \\
Obs & 224$\times$224 fisheye RGB + state \\
Action / horizon & 7-d / 8 steps \\
Optimizer & AdamW, lr=$10^{-4}$ \\
Batch / epochs & 192 / 200 \\
LR sched & Cosine, warmup 2000 steps \\
EMA & Power 0.75 \\
Flow ratio & 0.5 \\
$\alpha$ & 0.15 \\
Train / Infer & RTX 5090 / Jetson AGX Thor \\
\bottomrule
\end{tabular}
\end{minipage}
\end{table*}

\textbf{Onboard latency and closed-loop response.}
On the same Thor, DDIM-16 requires 152\,ms per action chunk versus 19\,ms for HybridFlow. Its sequential denoising thus adds 133\,ms before an updated prediction becomes available. The controller continues executing the previously issued trajectory while the next chunk is computed, including in static pick-and-place. Inference therefore overlaps physical execution; the reported evaluation does not pause the arm for each prediction. Longer inference delays the availability of feedback corrections. More denoising steps consume a larger share of the response budget on this edge platform; higher offline generation quality alone does not ensure better closed-loop task success. HybridFlow combines low latency with accurate actions under this shared deployment protocol.

The eightfold speedup measures action-generation latency with observation features already available: camera acquisition, visual encoding, and robot execution are excluded. Auxiliary camera-path measurements are approximately 95\,ms on Thor and 62\,ms on RTX 5090; all physical rollouts use the Thor path. Training and simulation hardware are listed separately in Table~\ref{tab:hyper}.

\textbf{Execution behavior.}
Plain MeanFlow fails to generate sufficiently accurate actions for the physical tasks evaluated here. Its predicted actions drove the arm beyond permitted motion limits, causing execution to halt before task completion; ReFlow exhibited the same failure mode. These are action-quality failures despite the policies' low inference latency. Extending real-robot MeanFlow training from the default 200 to 1,000 epochs yields only a small validation-MSE decrease, from 0.0065 to 0.0061, and does not resolve the poor action quality. This duration study is separate from the 2,000-epoch simulation training. DDIM-2 also exhibited grasp-offset errors.

Plain MeanFlow achieves 78\% in the simulation ablation and 0\% in the physical evaluation. These results come from separately trained policies with different observations, backbones, and task distributions, rather than direct sim-to-real transfer. The physical failures show that the generated actions lack the accuracy needed for execution in the evaluated UMI settings.

\subsection{VLA Action-Expert Transfer}
\label{sec:vla}
We also integrate HybridFlow as an action expert in StarVLA~\cite{starvla2026}. The vision-language backbone and connector remain frozen, while the original 4-step ReFlow action expert is replaced with a HybridFlow head using the same training recipe. Table~\ref{tab:vla} reports suite-level and aggregate results: on LIBERO, the 2-NFE expert remains competitive across Spatial, Object, and Goal; on RoboTwin~2.0, it reaches 89.3\% versus 88.3\% for the 4-step baseline, averaged over three seeds. The experiments demonstrate compatibility with the frozen vision-language backbone at half the action-expert evaluation count.

\subsection{Implementation Settings}
\label{sec:implementation}
Table~\ref{tab:hyper} summarizes training and network settings. Both calls in Algorithm~\ref{alg:hybridflow} reuse the same observation encoding; ReNoise is a tensor interpolation.

\FloatBarrier
\section{Limitations}
\label{sec:limitations}
HybridFlow is evaluated on the reported models and tasks; its empirically selected refinement ratio may require retuning for other action distributions. The propagation bound does not establish a causal link between query shift and failure. Plain MeanFlow's action-quality failures are documented in the tested UMI setting; their generality and response to other optimization or representation choices remain open. We do not separately quantify the contributions of action accuracy and feedback delay.

\section{Conclusion}
\label{sec:conclusion}
HybridFlow combines a full-interval MeanFlow proposal, time-aligned ReNoise, and diagonal instantaneous-velocity correction in a unified 2-NFE model. Controlled ablations support the proposal and refinement-state construction with either reused or fresh noise. With all real-robot policies on the same Jetson AGX Thor, it improves normalized task performance over DDIM-16 by 13--68 points with approximately eightfold lower action-generation latency, reaching 43/63 successes (68.3\%) in dynamic catching. Simulation and VLA experiments demonstrate broader applicability.

\FloatBarrier
\bibliographystyle{IEEEtran}
\bibliography{references}
\end{document}